\documentclass[sigconf]{acmart}
\usepackage{soul}
\usepackage{units}
\usepackage{algorithm}
\usepackage{algorithmic}
\usepackage{enumitem}
\usepackage{multirow}
\usepackage{subfigure}
\usepackage{bbm}
\usepackage{fdsymbol}

\newtheorem{theorem}{Theorem}
\newtheorem{definition}{Definition}
\newtheorem{problem}{Problem}

\usepackage{comment}

\AtBeginDocument{%
  \providecommand\BibTeX{{%
    \normalfont B\kern-0.5em{\scshape i\kern-0.25em b}\kern-0.8em\TeX}}}

\copyrightyear{2024}
\acmYear{2024}
\setcopyright{acmlicensed}
\acmConference[KDD '24] {Proceedings of the 30th ACM SIGKDD Conference on Knowledge Discovery and Data Mining }{August 25--29, 2024}{Barcelona, Spain.}
\acmBooktitle{Proceedings of the 30th ACM SIGKDD Conference on Knowledge Discovery and Data Mining (KDD '24), August 25--29, 2024, Barcelona, Spain}
\acmISBN{979-8-4007-0490-1/24/08}
\acmDOI{10.1145/3637528.3672050}
\begin{document}

\title{Motif-Consistent Counterfactuals with Adversarial Refinement for Graph-Level Anomaly Detection}

\author{Chunjing Xiao}
\affiliation{%
  \institution{Henan University}
  \country{China}
}
\email{chunjingxiao@gmail.com}

\author{Shikang Pang}
\affiliation{%
  \institution{Henan University}
  \country{China}
}
\email{pangsk0604@henu.edu.cn}

\author{Wenxin Tai}
\affiliation{%
  \institution{University of Electronic Science and Technology of China}
  \country{}
}
\email{wxtai@outlook.com}

\author{Yanlong Huang}
\affiliation{%
  \institution{University of Electronic Science and Technology of China}
  \country{}
}
\email{hyloong77@gmail.com}

\author{Goce Trajcevski}
\affiliation{%
  \institution{Iowa State University}
  \country{USA}
}
\email{gocet25@iastate.edu}

\author{Fan Zhou}
\authornote{Corresponding Author.}
\affiliation{%
  \institution{University of Electronic Science and Technology of China}
  \country{}
}
\email{fan.zhou@uestc.edu.cn}
\renewcommand{\shortauthors}{Chunjing Xiao et al.}
\begin{abstract}
  Graph-level anomaly detection is significant in diverse domains. To improve detection performance, counterfactual graphs have been exploited to benefit the generalization capacity by learning causal relations. Most existing studies  directly introduce perturbations (e.g., flipping edges) to generate counterfactual graphs, which are prone to alter the semantics of generated examples and make them off the data manifold, resulting in sub-optimal performance. To address these issues, we propose a novel approach, Motif-consistent Counterfactuals with Adversarial Refinement (MotifCAR), for graph-level anomaly detection. The model combines the motif of one graph, the core subgraph containing the identification (category) information, and the contextual subgraph (non-motif) of another graph to produce a raw counterfactual graph. However, the produced raw graph might be distorted and cannot satisfy the important counterfactual properties: \emph{Realism}, \emph{Validity}, \emph{Proximity} and \emph{Sparsity}. Towards that, we present a Generative Adversarial Network (GAN)-based graph optimizer to refine the raw counterfactual graphs. It adopts the discriminator to guide the generator to generate graphs close to realistic data, i.e., meet the property \emph{Realism}. Further, we design the motif consistency to force the motif of the generated graphs to be consistent with the realistic graphs, meeting the property \emph{Validity}. Also, we devise the contextual loss and connection loss to control the contextual subgraph and the newly added links to meet the properties \emph{Proximity} and \emph{Sparsity}. As a result, the model can generate high-quality counterfactual graphs. Experiments demonstrate the superiority of MotifCAR.
\end{abstract}





\maketitle

\newcommand{\tcr}{\textcolor{red}}
\newcommand{\tcb}{\textcolor{blue}}
\newcommand{\modelname}{MotifCAR}

\section{Introduction}
\label{sec:intro}




Graph-level anomaly detection aims to identify graph instances that are significantly different from the majority of graphs. As a few anomalies may cause tremendous loss, detecting anomalous data has significant implications for various domains ranging from 
identifying abnormal proteins in biochemistry and distinguishing brain disorders in brain networks, to uncovering fraudulent activities in online social networks
\cite{ma2021comprehensive,akoglu2015graph}. 
Numerous corresponding detection methods have been introduced 
by taking advantage of different deep learning techniques for anomaly detection\cite{ma2021comprehensive,xiao2023imputation}, such as self-supervised learning~\cite{qiu2022raising,zhao2023using}, knowledge distillation~\cite{ma2022deep,lin2023discriminative} and tailored Graph Neural Networks (GNNs)~\cite{qiu2022raising,zhang2022dual}.

However, these deep learning models are prone to learn dataset-dependent spurious correlations based on statistical associations~\cite{krueger2021out}. This might hinder well-trained models from generalizing well to newly observed anomalies, resulting in detection errors.
Counterfactual data augmentation can help the model alleviate the problem of spurious correlations by learning causal relations and enhance the generalization capacity in tabular data, image data, and text data~\cite{khorram2022cycle,chou2022counterfactuals}. 
While, for graph data, research on counterfactual augmentation is insufficient due to the presence of complex structure and node information.
The limited existing research principally focuses on introducing perturbations into graphs or matching counterfactual data with different treatments~\cite{ma2022learning,xiao2023counterfactual,zhao2022learning,chang2023knowledge}.


\begin{figure}[!tb] 
\centering
\includegraphics[width=0.46\textwidth]{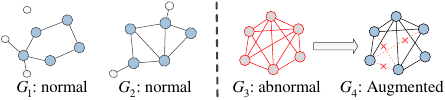}
\vspace{-0.35cm}
\caption{An example of perturbations altering semantics. $G_1$ and $G_2$ are normal graphs. While, $G_3$ is abnormal one as its fully connected structure deviates significantly from the normal ones. However, by pruning a few edges, abnormal graph $G_3$ is transformed into normal graph $G_4$, leading to an  altered semantic.}
\label{fig:CutEdgeProblem}
\vspace{-0.55cm}
\end{figure}

Whereas, when applying to anomaly detection, these methods confronts the  crucial problems:
(1) Perturbations might alter the graph semantics, adversely impacting model robustness. In anomaly detection, the category of a graph (e.g., normal or anomalous label) can be determined by the presence of specific edges~\cite{ma2021comprehensive}. Perturbing these edges may change its category, as illustrated in Figure~\ref{fig:CutEdgeProblem}.
However, the augmented samples will be labeled with the same category as the original ones~\cite{ma2022learning,xiao2023counterfactual,ding2022data}, resulting in incorrect labels and degraded performance.
(2) These methods may generate suboptimal augmented data, leading to limited effectiveness in enhancing generalization capacity. Perturbation methods often lead to distorted examples~\cite{szegedy2014intriguing}, which are usually off the data manifold~\cite{khorram2022cycle}. In such cases, deep models can be deceived because they do not generalize well to unseen test data~\cite{khorram2022cycle}. Also, matching approaches only seek desired samples from existing data~\cite{zhao2022learning,chang2023knowledge}, which cannot expand the training data.





In this paper, we propose a Motif-consistent Counterfactual data generation model with Adversarial Refinement (\modelname) to generate high-quality counterfactual samples for graph-level anomaly detection. 
In this model, to address the problem (1), instead of conducting perturbations on graphs, we introduce the \textit{discriminative motif} for counterfactual data generation. According to~\cite{han2022g}, the discriminative motif is a subgraph of a graph that decides the category of this graph.
Hence, it can be regarded as the core subgraph containing the identification (category) information of 
a given graph. 
Correspondingly, the remaining nodes can form a subgraph as the contextual graph.
For brevity, we refer to the discriminative motif simply as motif. 
Based on this concept, we combine the motif of one graph and the contextual subgraph of another graph to form a \textit{raw counterfactual graph}, i.e., unseen combinations of the motif and the contextual subgraph. These two source graphs can be from the same/different categories/clusters.

To address the problem (2), we design a Generative Adversarial Network (GAN)-based graph optimizer along with our tailored losses to refine the raw counterfactual graphs into the high-quality ones. 
According to~\cite{chou2022counterfactuals,moraffah2020causal}, good counterfactual data requires satisfying a set of properties (a.k.a. \textit{counterfactual properties}), including 
(1) \emph{Realism}: the counterfactual examples should lie close to the data manifold so that they appear realistic.
(2) \emph{Validity}: the model should assign the counterfactual examples to the corresponding category label in order to be valid;
(3) \emph{Proximity}: the distance of a counterfactual and original data should be close; 
(4) \emph{Sparsity}: the number of perturbations on nodes/edges should be sparse.
To meet the desired counterfactual properties, we design three specific losses for the graph optimizer.
The refined counterfactual graphs can enlarge the distribution of training data and help the model handle the situation with varying environments.

More succinctly, 
our \modelname~ model is composed of two fundamental components: the raw counterfactual graph producer and the GAN-based graph optimizer. 
The graph producer takes as inputs two graphs, $G$ and $H$, and produces a raw counterfactual graph by combining the motif of $G$ and the contextual subgraph (non-motif) of $H$.
This producer first merges these two graphs into one graph, and then discards the contextual subgraph of $G$ and the motif of $H$. The remaining parts (the motif of $G$ and the contextual subgraph of $H$) are connected randomly to form a raw counterfactual graph. The generated graph preserves the identification (category) information of $G$ but involves the environment characteristics of $H$. 
These unseen combinations can enlarge the distribution of training data and help the model learn transferable relations across different environments.
However, the random links between both subgraphs in the raw counterfactual graphs may distort the graph structure, which does not adhere well to the desired counterfactual properties.

Hence, we further design the GAN-based graph optimizer to refine the raw graphs by adjusting the edges.
The optimizer is composed of the graph generator and the discriminator. The generator aims to adjust the edges of the graphs and the discriminator tries to distinguish generated graphs from realistic ones. This adversarial training will encourage the generated graphs to conform to the patterns of realistic graphs, i.e., meet the requirement of the property (1) \emph{Realism}. On the basis of the basic GAN framework, we further design three losses to improve the quality of the generated graphs.
Firstly, we propose a motif consistency loss to force the motif of the generated graphs to be consistent with the realistic graphs. Since the motif contains the identification information of the graphs, this loss ensures the identification information invariant with the realistic graphs, i.e., meets the property (2) \emph{Validity}. 
Secondly, we devise the contextual loss to keep the degree distribution of the generated contextual subgraphs to be similar to the realistic ones. Coupled with the motif consistency loss, these losses facilitate 
the proximity of the counterfactual graphs to the original ones, i.e., the property (3) \emph{Proximity}.
Thirdly, we present the connection loss to control the new links between the motif and the contextual subgraph which, in turn, can control the number of perturbations, i.e., meet the property (4) \emph{Sparsity}.
Through the 
adversarial training with these losses, the model can generate counterfactual graphs that conform to the counterfactual properties. Finally, 
these generated counterfactual graphs and realistic graphs are adopted to train a robust anomaly detection model.
The contributions of this study are threefold:
\begin{itemize}[leftmargin=*]
\item We propose a new framework for graph-level anomaly detection, \modelname, 
which introduces a novel paradigm for generating counterfactual graphs using the motif.
\item We present the GAN-based graph optimizer, that develops the three tailored losses to meet the requirement of desired counterfactual properties and generate high-quality samples.
\item Extensive experiments demonstrate that \modelname~ significantly improves the detection performance and counterfactual data quality compared to state-of-the-art baselines. 
\end{itemize}






\section{Preliminaries}

We now provide a preliminary overview of the graph-level anomaly detection problem, the graphon, and the discriminative motif, 
which will serve as important foundational elements in our  model.


\subsection{Problem Formulation}


In general, a graph can be represented as $G(\mathcal{V},\mathcal{E})$, where $\mathcal{V}$ and $\mathcal{E}$ denote its node set and edge set, respectively. We adopt $G,H,I/\mathcal{G},\mathcal{H},\mathcal{I}$ to denote graphs/graph set. $\mathbf{y}_{G}\in\mathbb{R}^C$ denotes the label of graph $G$, where $C$ is number of classes of graphs. Accordingly, the problem of graph-level anomaly detection is defined as follows:
\begin{problem}
Graph-level anomaly detection aims to identify individual graphs within a given set $\mathcal{G}$ that exhibit anomalous behavior. In this study, the problem of graph-level anomaly detection can be regarded as the task of identifying anomalous graphs $G_i$ based on a limited set of graph labels $y_i = \{0,1\}$, where 0 signifies that the graph is considered an anomaly.
\label{def:problem}
\end{problem}

\subsection{Graphon}

We will leverage  grahpons to produce  counterfactual graphs. Here we present an introduction to the graphon and estimation method.

\noindent\textbf{Graphon}. 
A graphon \cite{airoldi2013stochastic} is a continuous, bounded and symmetric function
$W:[0,1]^{2}\rightarrow[0,1]$
which may be thought of as the weight matrix of a graph with infinite number of nodes. Then, given two points $v_{i},v_{j}\in[0,1], W(i,j)$ represents the probability that nodes $i$ and $j$ are related with an edge. 
For a set of graphs $\mathcal{G}$ with a given category label, a graphon can be estimated based on these graphs. 
Conversely, arbitrarily sized graphs can be sampled from this graphon, and sampled graphs will preserve the same category label with this graphon \cite{han2022g}.

\noindent \textbf{Graphon Estimation.} 
Give a graph set $\mathcal{G} = \{G_1, G_2, \ldots, G_n\}$, graphon estimation aims to deduce a graphon $W_G$ based on $\mathcal{G}$. 
It is intractable because a graphon is an unknown function without a closed-form expression for real-world graphs~\cite{han2022g}.
Hence, the step function estimation methods are generally adopted to approximate graphons~\cite{xu2021learning, han2022g}.
The typical step function estimation method is composed of two stages: aligning the nodes in a set of graphs and estimating the step function from all the aligned adjacency matrices~\cite{ han2022g}.
For the node alignment, this method first aligns multiple graphs based on node measurements (e.g., degree), and then selects the top $K$ nodes from the aligned graphs and calculates the average matrix of their adjacency matrices.
For function estimation, the goal is to obtain a matrix $\mathbf{W}=[w_{kk^{\prime}}]\in[0,1]^{K\times K}$, where $w_{ij}$ denotes the probability of an edge existing between node $i$ and node $j$ and $K$ is the number of the selected nodes, with the default being the average number of nodes in all graphs\cite{han2022g,xu2021learning}.
In particular, the step function $\mathbf{W}^{P}:[0,1]^{2}\mapsto[0,1]$ is defined as follows:
\begin{equation} 
\begin{aligned}
\mathbf{W}^{P}(x,y)=\sum_{k,k^{\prime}=1}^{K}w_{kk^{\prime}}
\mathbbm{1}_
{{P}_{k}
\times{P}_{k^{\prime}}}
(x,y),
\label{equ:graphonequ}
\end{aligned}
\end{equation}
$\mathcal{P} = ({P}_{1},...,{P}_{K})$ represents the partition of $[0,1]$ into $K$ adjacent intervals of length $1/K$, $w_{kk^{\prime}}\in[0,1]$, and the indicator function $\mathbbm{1}_{{P}_k\times{P}_{k^{ 
 \prime}}}(x,y)$ equals 1 if $(x,y)\in{P}_k\times{P}_{k^{\prime}}$ and 0 otherwise. 

\subsection{Discriminative Motif}
\label{sec:motifGraphon}


Motif, also called interest or frequent subgraph, refers to a specific subgraph structure that frequently occurs in a graph such as edges, triangles and quadrilaterals \cite{yuan2023motif}.
On the basis of the motif concept, similar to the work~\cite{han2022g}, we define the discriminative motif as:
\begin{definition}
A discriminative motif $F_G$ of the graph $G$ is the subgraph, which can decide the category of the graph $G$.
\label{def:definiMotif}
\end{definition}
Intuitively, the discriminative motif is the core subgraph of a graph and every graph has a discriminative motif, which generally has a smaller number of nodes and edges.
For the relationship of discriminative motifs and graphons, 
we have the theorem:
\begin{theorem}
For a graph $G$ which is sampled from the graphon $W_G$, its discriminative motif exists in $W_G$.
\label{def:theoremMotif}
\end{theorem}

\emph{Proof Sketch}. We verify this by stating that the homomorphism density of the discriminative motif in generated (sampled) graphs will be approximately equal to that in the graphon with high probability. 
In other words, the sampled graphs will preserve the discriminative motif of the graphon with a very high
probability.
Here, homomorphism density is defined to measure the relative frequency that the graph $H$ appears in graph $G$~\cite{lovasz2006limits,han2022g}.


\section{Method}
\label{sec:method}

\begin{figure}[!tb] 
\centering
\includegraphics[width=0.47\textwidth]{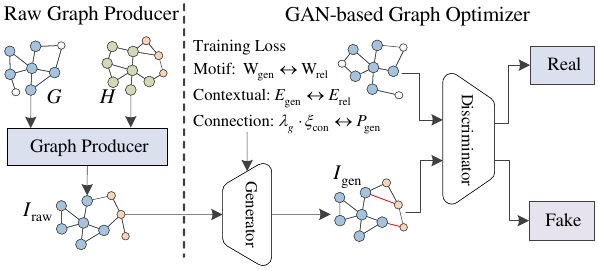}
\vspace{-0.3cm}
\caption{The Overview of the \modelname~ framework.}
\label{fig:FigMotiOverview}
\vspace{-0.40cm}
\end{figure}

As mentioned in Section~\ref{sec:intro}, \modelname~ 
is 
composed of two principal components: the raw counterfactual graph producer and the GAN-based graph optimizer, which are illustrated 
in Figure~\ref{fig:FigMotiOverview}. In the rest of this section, we discuss the technical details of \modelname.

\subsection{Producing Raw Counterfactual Graph}


Here we present the raw graph producer, which takes two graphs as inputs and extracts the motif subgraph from one graph and contextual subgraph from another to generate a raw counterfactual graph.
The generated graph possesses the label of the first graph while involving the environment information of the second graph. 
To this end, this producer first merges the two graphs into one graph, and then builds a masked matrix based on graphons of the categories of these two graphs to filter the motif and the contextual subgraph. Finally, the raw counterfactual graphs are produced by combining the adjacency matrix of the merged graph and the masked matrix. 
We adopt this generation way because: (1) this fixed merging pattern ensures a stable outcome for each merging attempt, and (2) it enables uniform expansion of the node counts when the two graphs have different numbers of nodes.

Formally, suppose that a graph ${G}$ with $n_g$ nodes is from the graph set $\mathcal{G}$ and graph ${H}$ with $n_h$ nodes from the graph set $\mathcal{H}$. Here ${G}$ and ${H}$ can have the same/different cluster or category label and their graphons can be computed using Equation~\ref{equ:graphonequ}, denoted as $W_G$ and $W_H$, individually.
To produce raw counterfactual graphs, we first merge two graphs ${G}$ and ${H}$ into one. 
Since ${G}$ and ${H}$ might have different numbers of nodes, we extend their node sets to a set $V_G \cup V_H$, and their adjacency matrices can be computed as:
\begin{equation} \begin{aligned}
A_{{G}}^{\mathrm{ext}}=
    \left[\begin{array}{cc}
        {A_{{G}}} & {0} \\
        {0}              & {0_{{H}}}\\
    \end{array}\right], 
A_{{H}}^{\mathrm{ext}}=
    \left[\begin{array}{cc}
        {0_{{G}}} & {0} \\
        {0}              & {A_{{H}}}\\
    \end{array}\right],
\label{equ:adjacencyExtend}
\end{aligned} \end{equation}
where $0_{G}$ and $0_{H}$ are zero matrices with shapes $n_g \times n_g$ and $n_h\times n_h$, respectively. 
Then, the two adjacency matrices are merged into $A_{p}$:
\begin{equation} \begin{aligned}
    A_{p} = A_{{G}}^{\mathrm{ext}} + A_{{H}}^{\mathrm{ext}}=\left[\begin{array}{cc}A_{{G}} & A_C\\A_C^T & A_{{H}} \end{array} \right],
\label{equ:AdjGrep}
\end{aligned} \end{equation}
where $A_C$ is a matrix indicating the cross-graph connectivity between the nodes in ${G}$ and ${H}$. $A_{{G}}$ and $A_{{H}}$ are the aligned adjacency matrices, i.e., their first $|W_G|$ and $|W_H|$ elements are aligned with their corresponding graphon nodes. We randomly sample $\eta$ edges to connect ${G}$ and ${H}$ to ensure the produced graph is connected.

To remove the contextual (non-motif) nodes in $G$ and remove the motif nodes in $H$, we build a mask matrix which is used to perform an XNOR operation with $A_{p}$ to eliminate these nodes. Since the motifs of $G$ and $H$ exist in the graphons $W_G$ and $W_H$, individually, we build this mask matrix based on $W_G$ and $W_H$. To this end, we first extend row and column numbers of $W_G$ and $W_H$ to be the same with $A_{{G}}$ and $A_{{H}}$, respectively:
\begin{equation} \begin{aligned}
 W_G^{\text{ext}}=
    \left[\begin{array}{cc}
        {W}_{G}          & 0 \\
        {0}              & 0_{\overline{G}}\\
    \end{array}\right], 
 W_{{H}}^{\text{ext}}=
    \left[\begin{array}{cc}
        {{W}_{{H}}}      & 0 \\
        {0}              & 0_{\overline{H}}\\
    \end{array}\right],
\label{equ:grphonExtend}
\end{aligned} \end{equation}
where $0_{\overline{G}}$ and $0_{\overline{H}}$ are zero matrices with shapes $n_{\overline{g}} \times n_{\overline{g}}$ and $n_{\overline{h}} \times n_{\overline{h}}$, individually. Here $n_{\overline{g}} = \big|| A_G| - |{W}_{G}| \big|$ and $n_{\overline{h}}=\big|| A_H| - |{W}_{H}|\big|$. Then, we build this mask matrix as:
\begin{equation} \begin{aligned}
   W^{m} = 
   \left| \begin{array}{cc} 
      \hat{W}_{G}^\text{ext}   & A_C\\
      A_C^T                     & A_H - \hat{W}_{H}^\text{ext}
   \end{array}\right|,
\label{equ:edgeMaskequ}
\end{aligned} \end{equation}
where $A_C$ is the same with that in Equation~\ref{equ:AdjGrep}. $\hat{W}_{G}^\text{ext}$ and $\hat{W}_{H}^\text{ext}$ are the matrices obtained after binary conversion of $W_{G}^\text{ext}$ and $W_{H}^\text{ext}$:
\begin{equation} \begin{aligned}
\hat{W}_{G}^\text{ext} =  Bern(W_{G}^\text{ext}), 
\hat{W}_{H}^\text{ext} =  Bern(W_{H}^\text{ext}),
\label{equ:maskequ}
\end{aligned} \end{equation}   
where $Bern(\cdot)$ refers to the Bernoulli function which binarizes the values of the matrix.   
As a result, we conduct the XNOR operation between $W^{m}$ and $A_p$ to obtain the adjacency matrix of the counterfactual graph:
\begin{equation} \begin{aligned}
A_{p}^m =  W^{m} \odot A_{p}.
\label{equ:xnor}
\end{aligned} \end{equation}
According to the adjacency matrix $A_{p}^m$, we can obtain the raw counterfactual graph $I_\text{raw}(\mathcal{V}, \mathcal{E})$, where $\mathcal{E}$ is derived from $A_{p}^m $ and $\mathcal{V}$ is composed of the motif nodes of $G$ and contextual nodes of $H$.

Since $I_\text{raw}(\mathcal{V}, \mathcal{E})$ is produced based on the motif of one graph and the contextual subgraph of another graph, it should possess the identification of the first graph and also involve the environment characteristics of the second graph. However, the motif nodes and the contextual nodes are connected randomly, which may result in distorted links. Hence, we further feed the produced raw counterfactual graphs into the graph optimizer to refine the graphs.

\subsection{Optimizing Counterfactual Graphs }


In the raw counterfactual graph ${I}_\text{raw}(\mathcal{V}, \mathcal{E})$, the edges between the motif nodes and the contextual nodes (i.e., $A_C$ in Equation \ref{equ:AdjGrep}) are randomly assigned, which might not conform to the distribution of realistic graphs. 
To obtain high-quality graphs, we here present a GAN-based graph optimizer to refine the raw graphs to meet the four counterfactual properties (\emph{Realism}, \emph{Validity}, \emph{Proximity} and \emph{Sparsity}). This optimizer is composed of two essential modules: the graph generator and the graph discriminator. The generator aims to generate refined graphs through adjusting edges. The discriminator is designed to distinguish between the graphs generated by the generator and realistic graphs. The graph generator and the graph discriminator are trained with our designed losses in an adversarial style to generate high-quality counterfactual graphs. These generated graphs will be used to train a robust classification model for better anomaly detection.

\subsubsection{Graph Generator.}

For the raw graph ${I}_\text{raw}(\mathcal{V}, \mathcal{E})$, the graph generator is designed to refine the edges in order to make the generated graph more closely resemble the real graph. We assume that each edge $\mathcal{E}(v_i, v_j)$  in ${I}_\text{raw}$ is associated with a random variable $P_{i,j} \sim Bern(\mathcal{W})$, where $ \mathcal{W} \in \mathbb{R}^{|\mathcal{V}| \times |\mathcal{V}|}$ is a learnable matrix, $P$ is a binary matrix with size $|\mathcal{V}| \times |\mathcal{V}|$, $\mathcal{E}(v_i, v_j)$ is in $I_\text{raw}$ if  $P_{i,j} = 1$ and is dropped otherwise.

Inspired by the work~\cite{wu2023graph}, we relax the discrete $P_{i,j}$ to a continuous variable with values in the range $(0,1)$ to facilitate end-to-end training of the generator:
\begin{equation} \begin{aligned}
    P=\sigma(\frac{\mathcal{W}-X_g}{\tau_g}),
\label{equ:disedge}
\end{aligned} \end{equation}
where $X_g \in \mathbb{R}^{|\mathcal{V}|\times|\mathcal{V}|} $ represents a random matrix with each element sampled from a uniform distribution: $X_{i,j} \sim U(0,1)$. $\sigma(x)=\frac{1}{1+e^{-x}}$ is the Sigmoid function, and $\tau_{g}\in(0,1]$ is a hyper-parameter used to make $P_{i,j}$ approach either 0 or 1. Here, $P$ can be regarded as an approximation of the generated adjacency matrix.

\textbf{Motif Consistency Loss}. The purpose of this loss is to ensure that the identification information of the generated graph is consistent with that of the given realistic graph. Since the graphon contains the motif of the graph and the motif decides the identification information, we adopt the graphon to build this consistency loss, i.e., we force the graphon of the generated graphs to be close to that of the realistic graphs.
For the generated graph set $\mathcal{I}_\text{gen} = \{{I}_\text{gen}^1, {I}_\text{gen}^2, \cdots , {I}_\text{gen}^n \}$ and the realistic graph set $\mathcal{G}_\text{rel} = \{{G}_\text{rel}^1, {G}_\text{rel}^2, \cdots , {G}_\text{rel}^n \}$,
their graphons ${W}_\text{gen}$ and ${W}_\text{rel}$ can be computed by Equation~\ref{equ:graphonequ}. Then, the loss is defined as:
\begin{equation} \begin{aligned}
    \mathcal{L}_{\mathrm{motif}} &= \sum_{i,j}||{W}_\text{gen}(i,j) -{W}_\text{rel}(i,j)||^{2}_F.
\label{equ:graphonloss}
\end{aligned} \end{equation}
This loss forces the generated graph to have the same motif as the realistic graphs. 
Since the motif decides the category label, this loss encourages the generated graphs to meet the property \emph{Validity}.


\textbf{Contextual loss}.
This loss is proposed to force the contextual subgraph to be similar to the realistic contextual one. 
Since the generator mainly adjusts the edges, 
we consider the edge-related feature -- the degree distribution -- for this loss, i.e., we force the degree distributions of the generated and realistic contextual subgraphs to be similar.
We introduce the degree entropy $E$ to measure the degree distribution of the subgraph. 
Supposing that $d(k)$ represents the degree number of node $v_k$ in a generated contextual subgraph, the degree entropy is computed as:
\begin{equation} \begin{aligned}
    E_\text{gen} & = -\frac{1}{ln(n_c)} \cdot \sum_{k=0}^{n_c}\frac{d(k)}{\sum_{m=0}^{n_c}d(m)}ln\bigg(\frac{d(k)}{\sum_{m=0}^{n_c}d(m)}\bigg), 
\label{equ:entropyequ}
\end{aligned} \end{equation}
where $n_c$ refers to the number of nodes in this contextual subgraph. This entropy indicates the average level of the node degree. Both entropies of the generated and realistic contextual subgraphs should be similar. Hence, the contextual loss is defined as:
\begin{equation} \begin{aligned}    
    \mathcal{L}_{\mathrm{context}} & = \sum_{i=1}^{n} \bigg| E_\text{gen}^i- E_\text{rel}^i\bigg|,
\label{equ:entropyloss}
\end{aligned} \end{equation}
where $E_\text{rel}$ is the entropy of the realistic contextual subgraph and $n$ denotes the number of the subgraphs. Both the contextual loss and the motif consistency loss compel the generated contextual subgraph and the motif to approach the realistic ones, which can meet the property \emph{Proximity}.

\textbf{Connection loss}.
This objective is designed to control the number of edges connecting the motif nodes and the contextual nodes in the generated graph. 
Inspired by the edge dropout \cite{zhu2020Deep}, we set a ratio $\lambda_{g}$ and train $\mathcal{W}$ to generate graphs with $\lambda_{g}\cdot|\mathcal{E}_\text{con}|$ connection edges. Here $\mathcal{E}_\text{con}$ is the edge set connecting the motif nodes and the contextual nodes in the realistic graph. For a batch of generated graphs $\mathcal{I}_\text{gen}=\{I_\text{gen}^{1},I_\text{gen}^{2},...,I_\text{gen}^{n}\}$, we calculate their average as the connection loss. Formally, this loss is computed as:
 \begin{equation} \begin{aligned}
    \mathcal{L}_\text{con} = 
    \frac{1}{n} \sum_{k=1}^{n} 
    \Big| \lambda_{g}\cdot| \mathcal{E}_\text{con}^k|
    - P^k_\text{gen} \Big|,
\label{equ:connectionloss}
\end{aligned} \end{equation}
where the $P_\text{gen}$ represents the sum of $P$ in the connection edge set of the generated graph, which is computed using Equation~\ref{equ:disedge}. This loss tries to limit the new connection edges, which can control the property \emph{Sparsity} of the generated graphs.

As a result, the regularization loss is the combination of the above losses:
\begin{equation} \begin{aligned}    
  \mathcal{L}_\text{reg}= 
      \lambda_{1} \cdot \mathcal{L}_\text{motif}
     +\lambda_{2} \cdot \mathcal{L}_\text{context}     
     +\lambda_{3} \cdot \mathcal{L}_\text{con},
\label{equ:regularizationLoss}
\end{aligned} \end{equation}
where $\lambda_{1}$, $\lambda_{2}$ and $\lambda_{3}$ are hyper-parameters to balance the influences of these losses.

Moreover, for the sake of efficiency, we initialize $\mathcal{W}$ rather than training from scratch. To be specific, we introduce an initialization rate $\gamma\in [0,1]$ to constrain the number of the connection edges at the beginning. Also the similarity of nodes is considered for the initialized values.
Consequently, $\mathcal{W}$ is initialized as follows:
\begin{equation} \begin{aligned}
  \mathcal{W}_{i,j} = \left\{ \begin{array}{cccc}
   1                        & , & \mathrm{if}(v_i,v_j)\in \mathcal{E}_t; \\
   \frac{\gamma\cdot \lambda_g \cdot|\mathcal{E}_\text{con}| \cdot S_{i,j}}{|C|} 
                            & , & \mathrm{if}(v_i,v_j)\in\mathcal{C}; \\ 
   0                        & , & \mathrm{otherwise},\end{array}\right.
\label{equ:w_init}
\end{aligned} \end{equation}
where $\mathcal{E}_t$ is the edge set in the motif and the contextual subgraph,  $\mathcal{E}_\text{con}$ and $\lambda_g$ are the same to Equation~\ref{equ:connectionloss}, $S_{i,j}$ is the node similarity which can be computed based on node embedding or node attributes, and $C$ is a candidate set of new connection edges.

\subsubsection{Graph Discriminator.}

The discriminator is a graph-level classifier designed to distinguish between real and generated graphs. This can help the generator produce samples that are close to realistic data, i.e., meet the property \emph{Realism}.
Specifically, this discriminator takes a graph as input and determines whether it is real or fake. 
Supposing that $\mathcal{G}_\text{rel}$ and $\mathcal{I}_\text{gen}$ denote the realistic graph set and the generated graph set, individually, for each $G \in \mathcal{G}_\text{rel} \cup \mathcal{I}_\text{gen}$, we utilize a GNN encoder $f$ to encode the representations of each node:
\begin{equation} \begin{aligned}
   \{\textbf{z}_{v}|v\in\mathcal{V}\}=f(G).
\label{equ:disequ}
\end{aligned} \end{equation}
Then, we calculate the graph representation by concatenating the mean pooling and the maximum pooling of the node representations:
 \begin{equation} \begin{aligned}
\textbf{z}_G = 
    \Big(\frac{1}{| \mathcal{V}|} \sum_{v\in\mathcal{V}} \textbf{z}_v \Big) 
    \oplus 
    \mathrm{MaxPool} \Big( \{\textbf{z}_v|v\in\mathcal{V}\} \Big),
\label{equ:graphRep}
\end{aligned} \end{equation}
where $\oplus$ is the concatenate operation. With the graph representation, we compute the probability of $G$ using an $L_h$-layer Multilayer Perceptron (MLP) $h$:
 \begin{equation} \begin{aligned}
p_{G}=h(\textbf{z}_{G}).
\label{equ:probG}
\end{aligned} \end{equation}
To train the discriminator, we label the graphs in $\mathcal{G}_\text{rel}$ with 1 and the graphs in $\mathcal{I}_{gen}$  with 0. Suppose that the label of $G$ is $l_{G}$. Then, the classification loss is defined as follows:
\begin{equation} \begin{aligned}
    \mathcal{L}_\text{dis}=-l_{G}\cdot\log(p_{G})-(1-l_{G})\cdot\log(1-p_{G}).
\label{equ:discriminatorLoss}
\end{aligned} \end{equation}

\subsubsection{Model Training.}
\label{GanTraining}


We here present the model training procedure of the GAN-based graph optimizer. 
The graph generator and the discriminator are optimized sequentially and iteratively.

For the generator, 
in each iteration, an augmented graph $I_\text{gen}$ is generated and then the regularization loss (cf. Equation~\ref{equ:regularizationLoss}) is computed. In consideration of generating high-quality graphs, an adversarial classification loss is incorporated to cheat the graph discriminator by labeling $I_\text{gen}$ with 1. According to the first term in Equation~\ref{equ:discriminatorLoss} and the regularization loss (cf. Equation~\ref{equ:regularizationLoss}), we have the final loss of the generator:
\begin{equation} \begin{aligned}
    \mathcal{L}_\text{gen}=-\log(p_{I_\text{gen}})-\mathcal{L}_{reg}.
\label{equ:generatorLoss}
\end{aligned} \end{equation}

For the discriminator, their parameters are optimized by classifying the realistic graphs and the generated graphs using the loss in Equation~\ref{equ:discriminatorLoss}.
For the computational complexity, since our MotifCAR is designed based on GANs, it has the similar complexity and overhead with GAN-based baselines (e.g., CFAD~\cite{xiao2023counterfactual}).



\subsection{Graph-level Anomaly Detection}


After training, the model is used to generate a number of counterfactual graphs. Both the generated counterfactual graphs $\mathcal{I}_\text{gen}$ and realistic graphs $\mathcal{G}_\text{rel}$ are adopted to train a robust classifier to identify anomalies from normal graphs. We adopt the discriminator as the classifier after changing its loss. Specifically, 
for each $G \in \mathcal{G}_\text{rel} \cup \mathcal{I}_\text{gen}$, we first adopt Equation~\ref{equ:probG} to obtain its graph representation. Then we label the normal graph with 1 and the anomalous graph with 0. Supposing that $y_G$ denotes the label, the loss becomes:
\begin{equation} \begin{aligned}
\mathcal{L}_{class}=-y_{G}\cdot\log(p_{G})-(1-y_{G})\cdot\log(1-p_{G}).
\label{equ:classAnomaly}
\end{aligned} \end{equation}
Moreover, to enhance detection performance, we refine the graph representations following the work~\cite{ma2023towards} during the process of the model training.

\begin{table}[t]
\small
\caption{Statistics of Datasets.} 
\label{table:datasets}
\vspace{-3mm}
\centering
\resizebox{0.47\textwidth}{!}{
\begin{tabular}{lrrrrrr}
\toprule
  Datasets  &\# graph &\# class & avg.\# V & avg.\# E & \ anomaly(\%)  \\
\midrule
  IMDB-B    &1,000    & 2       & 19.77    & 96.53	& 9.0\%  \\
  IMDB-M    &1,500    & 3       & 13.00    & 65.94	& 4.7\%  \\
  REDDIT-B  &2,000    & 2       &  429.63  &  497.75	& 9.0\%  \\
  REDDIT-M  &5,000    & 5       & 508.52   &  594.87  & 4.7\% \\
\bottomrule
\end{tabular}}
\vspace{-2.5mm}
\end{table}

\newcommand{\tabincell}[2]{\begin{tabular}{@{}#1@{}}#2\end{tabular}}
\begin{table*}[t]
\small
  \centering
  \caption{Performance comparison between \modelname~ and baselines on four datasets.}
  \label{tab:detectResults}%
 \resizebox{\textwidth}{!}{
    \begin{tabular}{@{}p{1.56cm}cccccccccccccc@{}}
\toprule
           \textbf{Method} & \multicolumn{3}{c}{\textbf{IMDB-BINARY}}              & \multicolumn{3}{c}{\textbf{IMDB-MULTI}}            & \multicolumn{3}{c}{\textbf{REDDIT-BINARY}}    & \multicolumn{3}{c}{\textbf{REDDIT-MULTI-5K}}            \\
\cmidrule(lr){2-4} \cmidrule(lr){5-7} \cmidrule(l){8-10} \cmidrule(l){11-13} 
           & \footnotesize{PRECISION } & \footnotesize{RECALL} & \footnotesize{F1}  & \footnotesize{PRECISION } & \footnotesize{RECALL} & \footnotesize{F1}  &   \footnotesize{PRECISION } & \footnotesize{RECALL} & \footnotesize{F1}  & \footnotesize{PRECISION } & \footnotesize{RECALL} & \footnotesize{F1}  &   \\
\midrule
        g-U-Nets & 0.87±0.01 & 0.91±0.01 & 0.88±0.01 & 0.80±0.01 & 0.86±0.01 & 0.80±0.01 & 0.86±0.01 & 0.75±0.02 & 0.78±0.01 & 0.76±0.01 & 0.71±0.02 & 0.73±0.01 \\
        DiffPool & 0.66±0.01 & 0.69±0.01 & 0.81±0.02 & 0.65±0.01 & 0.70±0.01 & 0.83±0.02 & 0.71±0.01 & 0.63±0.01 & 0.64±0.01 & 0.71±0.01 & 0.61±0.01 & 0.71±0.01 \\ 
        SAGPool & 0.80±0.01 & 0.88±0.01 & 0.84±0.01 & 0.73±0.01 & 0.90±0.01 & 0.82±0.01 & 0.81±0.01 & 0.81±0.01 & 0.83±0.01 & 0.76±0.01 & 0.61±0.01 & 0.74±0.01 \\
        GMT & 0.83±0.03 & 0.88±0.03 & 0.88±0.02 & 0.82±0.02 & 0.84±0.02 & 0.83±0.02 & 0.83±0.02 & 0.83±0.01 & 0.83±0.02 & 0.78±0.02 & 0.80±0.01 & 0.67±0.02 \\
        \midrule
        CFGL-LCR & 0.85±0.03 & 0.87±0.03 & 0.84±0.03 & 0.83±0.04 & 0.83±0.02 & 0.76±0.02 & 086±0.02 & 0.88±0.02 & 0.82±0.03 & 0.82±0.03 & 0.80±0.03 & 0.78±0.03 \\
        CFAD & 0.87±0.03 & 0.88±0.02 & 0.86±0.03 & 0.86±0.03 & 0.88±0.02 & 0.78±0.03 & 0.87±0.03 & 0.86±0.02 & 0.84±0.03 & 0.83±0.02 & 0.81±0.03 & 0.82±0.02 \\
        CGC & 0.92±0.02 & 0.94±0.03 & 0.92±0.02 & 0.89±0.02 & 0.93±0.02 & 0.92±0.02 & 0.90±0.03 & 0.93±0.02 & 0.91±0.02 & 0.84±0.03 & 0.95±0.02 & 0.84±0.02 \\
        CF-HGExp & 0.93±0.01 & 0.94±0.01 & 0.94±0.01 & 0.90±0.01 & 0.95±0.01 & 0.94±0.01 & 0.92±0.01 & 0.94±0.01 & 0.93±0.01 & 0.85±0.01 & 0.95±0.01 & 0.85±0.01 \\
        \midrule        
        OCGTL & 0.88±0.01 & 0.89±0.01 & 0.87±0.02 & 0.82±0.01 & 0.84±0.01 & 0.80±0.02 & 0.89±0.01 & 0.89±0.01 & 0.88±0.01 & 0.86±0.01 & 0.84±0.01 & 0.82±0.01 \\
        GLocalKD & 0.65±0.01 & 0.66±0.01 & 0.66±0.01 & 0.60±0.01 & 0.63±0.01 & 0.61±0.01 & 0.63±0.01 & 0.64±0.01 & 0.63±0.01 & 0.58±0.01 & 0.62±0.01 & 0.73±0.01 \\ 
        iGAD & 0.88±0.03 & 0.78±0.02 & 0.87±0.02 & 0.86±0.02 & 0.75±0.02 & 0.82±0.02 & 0.57±0.02 & 0.58±0.03 & 0.55±0.02 & 0.53±0.02 & 0.54±0.03 & 0.67±0.02 \\
        GmapAD & 0.92±0.02 & 0.94±0.01 & 0.94±0.01 & 0.90±0.01 & 0.95±0.01 & 0.93±0.01 & 0.91±0.02 & \textbf{0.96±0.01} & 0.95±0.01 & 0.87±0.02 & 0.95±0.01 & 0.86±0.01 \\
        \midrule

       \textbf{\modelname~} & \textbf{0.94±0.01} & \textbf{0.95±0.01} & \textbf{0.96±0.01} & \textbf{0.91±0.01} & \textbf{0.96±0.01} & \textbf{0.94±0.01} & \textbf{0.93±0.01} & \textbf{0.96±0.01} & \textbf{0.96±0.01} & \textbf{0.89±0.01} &  \textbf{0.97±0.01} & \textbf{0.87±0.01} \\ 

\bottomrule
    \end{tabular}
} 
\end{table*}

\section{Expriments}
\label{sec:expriment}

In this section, we evaluate the \modelname~ performance, including the effectiveness, ablation study, hyper-parameter analysis and the quality of the generated counterfactual graphs.

\subsection{Experiments Setup.}


\noindent\textbf{Datasets.} We adopt four public datasets for our experiments, whose statistics are presented in Table~\ref{table:datasets}.
\textit{IMDB-BINARY} (\textit{IMDB-B}) and \textit{IMDB-MULTI} (\textit{IMDB-M}) are movie collaboration datasets. Each graph corresponds to an ego-network for each actor/actress, where nodes correspond to actors/actresses and an edge is drawn between two actors/actresses if they appear in the same movie. Each graph is derived from a pre-specified genre of movies, which is regarded as the category label.
\textit{REDDIT-BINARY} (\textit{REDDIT-B}) and \textit{REDDIT-MULTI-5K} (\textit{REDDIT-M}) are balanced datasets where each graph corresponds to an online discussion thread and nodes correspond to users. An edge is drawn between two nodes if at least one of them responds to another's comment. The community or subreddit is considered as the label of the graph. 
Following previous works~\cite{qiu2022raising,ma2023towards}, we downsample one of the categories as the anomalous one, and others as the normal data. 



\noindent\textbf{Baselines}. We conduct a comparison between our developed framework \modelname~ and three categories of baselines: 
(1) The state-of-the-art GNN models (\textit{g-U-Nets} \cite{gao2019graph}, \textit{SAGPool} \cite{lee2019self}, \textit{DIFFPOOL} \cite{ying2018hierarchical}, and \textit{GMT} \cite{baek2021accurate}) leverage different pooling strategies and specially designed pooling layers for learning the graph-level representations.
(2) The counterfactual graph augmentation methods (\textit{CFGL-LCR}~\cite{zhang2023cfgl}, \textit{CFAD}~\cite{xiao2023counterfactual}, \textit{CGC}~\cite{yang2023generating}, and \textit{CF-HGExp}~\cite{yang2023counterfactual}) generate counterfactual graphs to enhance the node or graph representations and further improve the classification performance. 
(3) The graph-level anomaly detection approaches (\textit{OCGTL}~\cite{qiu2022raising}, \textit{GLocalKD}~\cite{ma2022deep},  \textit{iGAD}~\cite{zhang2022dual} and \textit{GmapAD}~\cite{ma2023towards}) explore the tailored classification models or graph-level anomaly patterns to 
the anomalies.


\noindent\noindent\textbf{Experiments Settings.}
We use the Adam algorithm to optimize the model with learning rate 0.001. 
For hyper-parameters, we set $\tau_g$ = 0.0001 in Equation~\ref{equ:disedge}, $\gamma$ = 0.75 in Equation~\ref{equ:w_init}, $\lambda_{1}$ = 1, $\lambda_{2}$ = 0.9, $\lambda_{3}$ = 0.6 in Equation~\ref{fig:hyperParameters}, $\lambda_{g}$ = 0.5 for IMDB-M and REDDIT-M and $\lambda_{g}$ = 0.8 for IMDB-B and REDDIT-B in Equation \ref{equ:connectionloss}.
We split the dataset into train/validation/test data by 2 : 4 : 4. 
The best test epoch is selected on the validation set, and we report the test accuracy on ten runs.
All the experiments are conducted on an Ubuntu 20.04 server with a 12-core CPU, 1 Nvidia RTX 3090 GPU and 64Gb RAM.

\subsection{Anomaly Detection Results}

We report the anomaly detection performance of \modelname~ and the baselines in Table \ref{tab:detectResults} and have the following observations.
First, 
in terms of overall detection results, our model \modelname~ consistently outperforms all baseline models across the four datasets. In particular, our method achieves more gains on IMDB-Binary and REDDIT-Binary. This is because graphon estimation can perform better on dense graphs~\cite{lovasz2012large} and the graphs in these two datasets are denser. 
Correspondingly, our model exhibits more significant advantages on these two datasets. 
This result verifies that our model can generate effective counterfactual graphs and enhance detection performance.
Besides, the $t$-test indicates that MotifCAR's performance is statistically significant compared to the baselines.

Second, 
some 
of the counterfactual graph augmentation methods, such as CGC and CF-HGExp, acquire relatively good performance, compared with the GNN models. This indicates that the appropriate counterfactual data can benefit graph-level anomaly detection. However, our model outperforms these state-of-the-art models, which validates the efficacy of the proposed GAN-based graph optimizer with the tailored losses for graph-level anomaly detection.

Third, the method GmapAD, which is specially designed for graph-level anomaly detection, exhibits remarkable performance. It can explore both the intra- and inter-graph node information to enhance the graph representations and detection performance. However, when faced with a smaller amount of training data, the trained model cannot generalize well to the test data, leading to a performance decline. However, our model can generate counterfactual data to handle varying environments and hence, outperform this method.


\subsection{Ablation Study}
\begin{figure*}[!tb]
\centering
\includegraphics[width=0.96\textwidth]{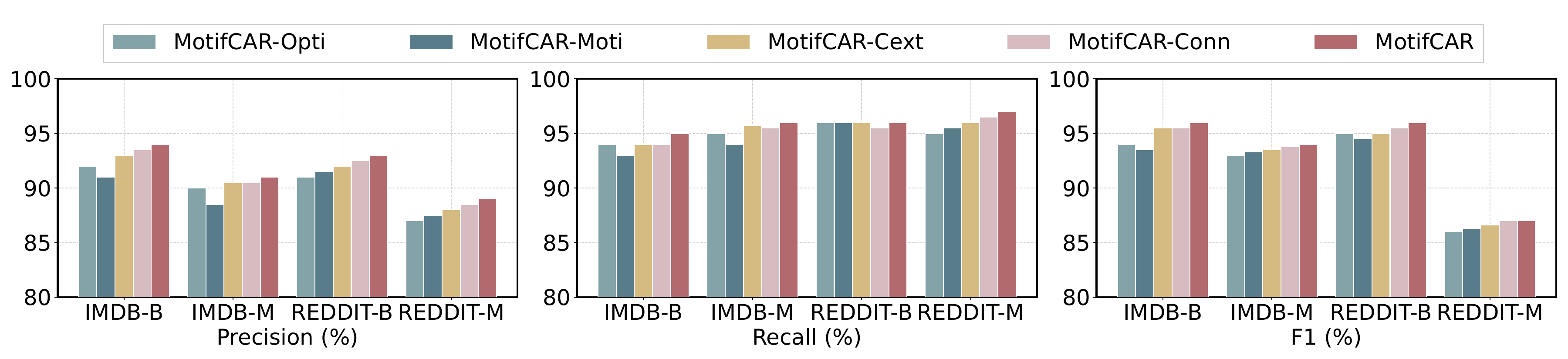}
\vspace{-3mm}
\caption{Ablation study: Variants of \modelname.}
\label{fig:Ablation}
\end{figure*}

In this section, we investigate the contributions of the key components in \modelname. We mainly observe the following variants: 
(1) In \textit{\modelname-Opti}, we remove the GAN-based graph optimizer.
(2) In \textit{\modelname-Moti}, we remove the motif consistency loss in the GAN-based optimizer.
(3) In \textit{\modelname-Cext}, we remove the contextual loss in the optimizer. 
(4) In \textit{\modelname-Conn}, we remove the connection loss in the optimizer. 

As shown in Figure~\ref{fig:Ablation}, no matter which part we remove in \modelname, the model's performance degrades. This suggests the efficacy of our designed GAN-based optimizer as well as the losses in this optimizer. 
Moreover, when removing the motif consistency loss, the performance of \textit{\modelname-Moti} delines more dramatically than \textit{\modelname-Cext} and \textit{\modelname-Conn}. This indicates that the motif consistency loss is more important when refining the counterfactual graphs. The reason is that this loss can ensure the generated graphs to have the corresponding identification information, which is vital for training a robust classification model.

\subsection{Qualitative Analysis of the Counterfactuals}


We here investigate the quality of the generated counterfactual graphs from the aspects of the counterfactual properties: \emph{Realism},  \emph{Proximity}, \emph{Validity}, and  \emph{Sparsity}. 
According to the works~\cite{khorram2022cycle,yang2023counterfactual}, the \emph{Realism} score reflects the change of the shift detection result between the realistic graphs and the counterfactual graphs. 
The \emph{Proximity} score estimates the mean of feature distances between the counterfactual graph and the realistic graph.
The \emph{Validity} score gauges the fraction of the generated counterfactual examples that are correctly predicted by the classifier to the corresponding class.
The \emph{Sparsity} score measures the difference between the edges of the counterfactual graph and the realistic graph.
We compare \modelname~ with  counterfactual data generation baselines: \textit{CFGL-LCR} (CFGL)~\cite{zhang2023cfgl}, \textit{CFAD} \cite{xiao2023counterfactual}, \textit{CGC}\cite{yang2023generating}, and \textit{CF-HGExp} (CF-HG)~\cite{yang2023counterfactual}). 

Table~\ref{tab:QuantResults} demonstrates the \emph{Realism} and \emph{Proximity} scores. 
For the Realism score (the lower the better), our method achieves the best result on the four datasets. 
The reason is that adversarial training is beneficial to generate counterfactual graphs that are close to realistic ones. Also, the motif consistency and contextual loss also facilitate aligning the motif and contextual subgraph well.
For the Proximity score (the lower the better), our model \modelname~ does not obtain the best result. This is because \modelname~ replaces the contextual subgraph in the counterfactual graphs, which results in a few differences in the feature space. While our result is competitive to the best baseline, CF-HGExp. In fact, both of their scores are very close. However, our model can achieve better detection performance than CF-HGExp.

\begin{table}[!ht]
        \centering
         \caption{Realism and Proximity.}
        \label{tab:QuantResults}
        \resizebox{0.47\textwidth}{!}{
        \begin{tabular}{l|l|ccccc}
        \toprule
        \multicolumn{2}{c|}{~} & CFGL & CFAD & CGC & CF-HG & \modelname~ \\
        \midrule
        \multirow{4}{*}{\rotatebox{90}{Realism}}& IMDB-B   
        & 1.980    & 1.84  & 0.844 & 0.688 & \textbf{0.622} \\
        & IMDB-M   & 1.946 & 1.850 & 0.752 & 0.562& \textbf{0.514} \\ 
        & REDDIT-B & 1.910 & 1.844 & 1.066 & 0.850& \textbf{0.784} \\ 
        & REDDIT-M & 1.896 & 1.860 & 1.142 & 0.754 & \textbf{0.462} \\ 
        \midrule
        \midrule
        \multirow{4}{*}{\rotatebox{90}{Proximity}}
        & IMDB-B   & 0.532 & 0.519 & 0.179 & 0.098          & \textbf{0.076} \\ 
        & IMDB-M   & 0.548 & 0.558 & 0.118 & \textbf{0.075} & 0.078 \\ 
        & REDDIT-B & 0.613 & 0.578 & 0.189 & \textbf{0.057} & 0.065 \\ 
        & REDDIT-M & 0.656 & 0.623 & 0.185 & \textbf{0.129} & 0.149 \\ 
        \bottomrule
        \end{tabular}}
        \vspace{-0.3cm}
\end{table}

\begin{figure}[h!] 
\centering
\vspace{-0.3cm}
\includegraphics[width=0.47\textwidth]{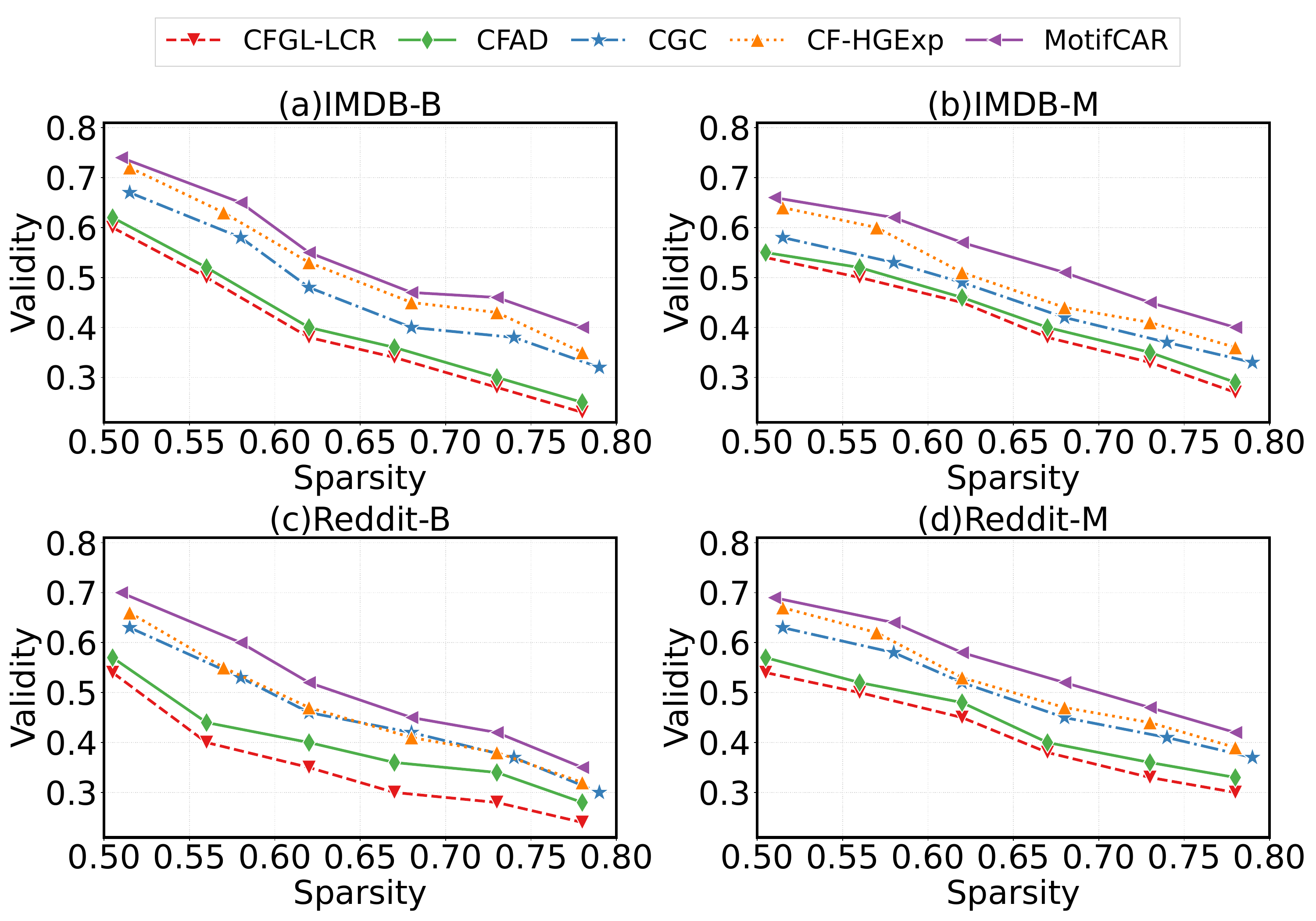}
\vspace{-0.2cm}
\caption{Validity and Sparsity.}
\label{fig:realism}
\vspace{-0.4cm}
\end{figure}

The \emph{Validity} and \emph{Sparsity} scores are presented in Figure~\ref{fig:realism},
where the misalignment of points on the x-axis is caused by discontinuous Sparsity values.  
As shown, our proposed \modelname~ achieves the best Validity performance at all levels of sparsity on the four datasets. 
This is because the raw counterfactual graphs contain the motif, the core subgraph of the graph, which can decide their category. Further, during the graph refinement procedure, the motif consistency provides strong consistent power for the counterfactual graphs keeping the identification information. As a result, the generated counterfactual graphs can possess the identification information and obtain higher Validity scores.



\subsection{Hyper-Parameter Sensitivity}

\begin{figure}[h!] 
\centering
\includegraphics[width=0.47\textwidth]{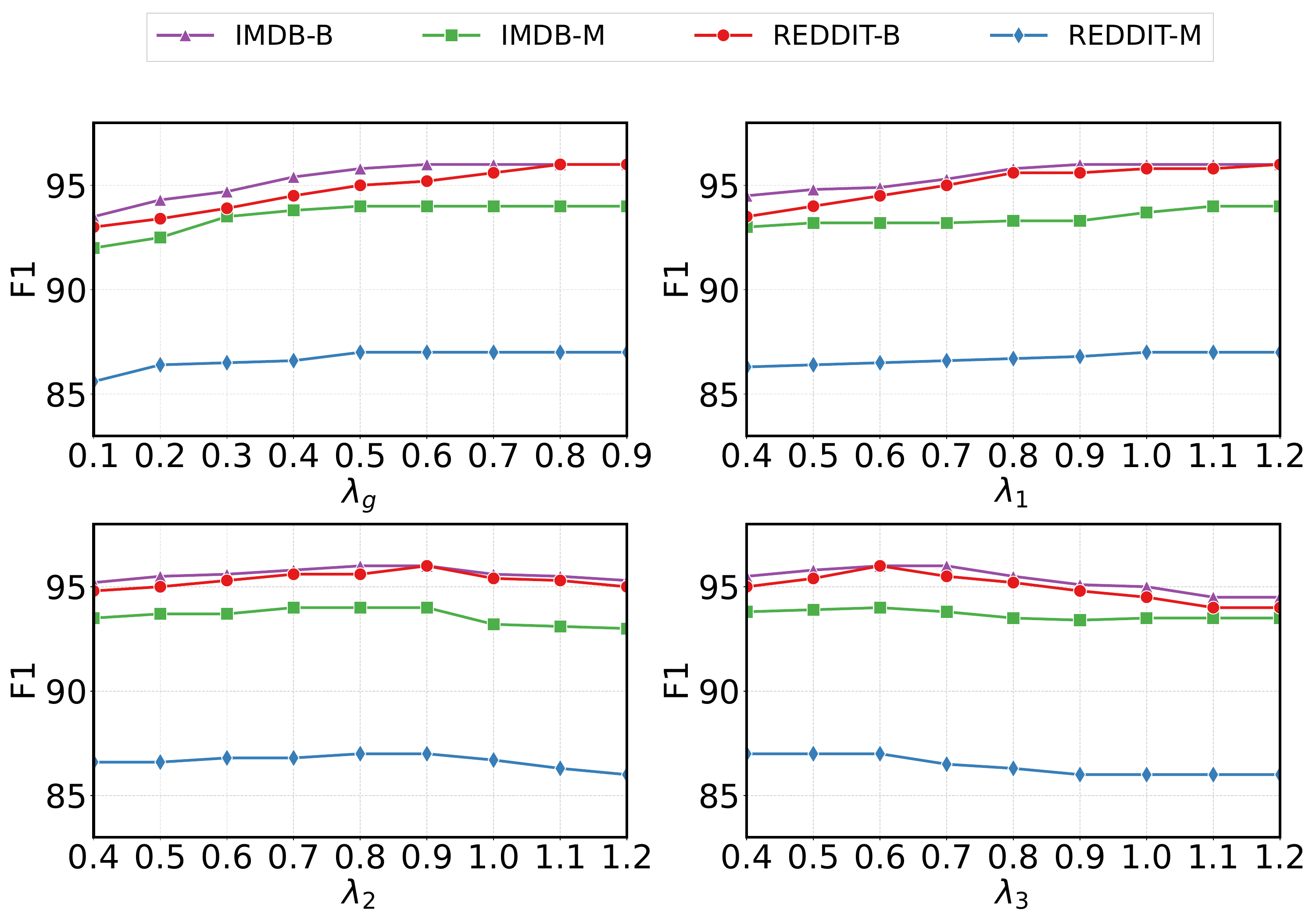}
\vspace{-0.2cm}
\caption{The effect of the hyper-parameters.}
\label{fig:hyperParameters}
\vspace{-0.45cm}
\end{figure}

In this section, we analyze the sensitivity of the  hyper-parameters, $\lambda_{g}$,
$\lambda_{1}$, $\lambda_{2}$, and $\lambda_{3}$ in Equation \ref{equ:connectionloss} and \ref{equ:regularizationLoss}. The results are reported in Figure~\ref{fig:hyperParameters}. For $\lambda_{g}$, the performance increases with the rise of the $\lambda_{g}$ value. A smaller value leads to a lower F1-score. This is because $\lambda_{g}$ determines the number of edges between the motif and the contextual subgraph, and a smaller value might cause the two subgraphs to disconnect. Correspondingly, the generated counterfactual graphs mainly contain the motif, which has little effect on improving the model's generalization ability. 

For $\lambda_{1}$, $\lambda_{2}$, and $\lambda_{3}$, these three parameters determine the weights of the motif consistency loss, the contextual loss, and the connection loss, individually. 
These figures show that F1-score increases with the rise of $\lambda_{1}$, which suggests that more attention should be paid to the motif consistency loss. The reason is that this loss determines the identification information of the generated counterfactual samples, which is quite important for training a robust model.
Besides, the detection performance increases with the rise of $\lambda_{2}$ and $\lambda_{3}$, and then starts to decrease when $\lambda_{2} > 0.9$ and $\lambda_{3} > 0.6$. When they are too small, the contextual subgraph cannot effectively be incorporated into the generated counterfactual graphs, resulting in a lower performance.
On the contrary, the higher $\lambda_{2}$ and $\lambda_{3}$ might attenuate the effectiveness of other losses, leading to inferior performance.

\section{Related Work}

\label{sec:relate}

\textbf{Graph-level Anomaly Detection.} 
Graph-level Anomaly Detection focuses on detecting entire abnormal graphs, rather than localizing anomalies in graphs. Most graph anomaly detection work is devoted to detecting irregular nodes and edges within graphs~\cite{ma2021comprehensive,duan2023graph,li2023controlled,tang2022rethinking}. Recently, graph-level anomaly detection has started to receive in-depth exploration. 
Initially, researchers adopt shallow learning techniques to detect graph-level anomalies. They
mainly exploit graph kernels and graph signals to detect graph-level anomalies. For example, graph kernels (e.g., Weisfeiler-Leman Kernel and propagation kernels) are explored to measure the pairwise similarity of nodes in the graph, and the similarity score is used to identify anomalies based on the structural characteristics of the graph \cite{manzoor2016fast}. Changedar \cite{hooi2018changedar} outlines anomaly detection using graph signals generated by anomalous node sets. By examining patterns of signal changes, this approach can effectively pinpoint anomalies within the graph. Also, a number of works \cite{noble2003graph, yuan2023motif, huang2021hybrid} leverage frequent graph motifs to model the network topology among nodes and motifs. These frequently occurring subgraphs capture crucial high-order structural information, which enables models to detect anomalies more comprehensively. However, the shallow learning-based methods may achieve sub-optimal performance due to the low coupling of the detector and graph representation learning~\cite{niu2023graph}.

The deep learning-based methods, in contrast, are end-to-end strategies that have been successfully applied to both static and dynamic graphs for anomaly detection~\cite{ding2019deep, yoon2019fast, zheng2019addgraph}. The remarkable development of GNNs~\cite{LAGCN,chen2024macro} has led to great progress in the field of graph-level anomaly detection. One approach involves utilizing GNNs in conjunction with classification loss functions to train a graph-level anomaly detection framework, exemplified by the works such as \textit{OCGIN}~\cite{zhao2023using} and \textit{OCGTL}~\cite{qiu2022raising}. However, these classification methods do not cope well with imbalanced datasets, resulting in an underfit for anomalous graphs. \textit{iGAD}~\cite{zhang2022dual} tackles this problem by modeling Point Mutual Information. And the out-of-distribution problem of graph data is addressed for better anomaly detection \cite{li2022graphde, liu2023good}.
Another method centers around the identification of anomalies by scrutinizing the irregular attributes within each graph concerning the overall graph structure, as evidenced in \textit{GmapAD}~\cite{ma2023towards} and \textit{GLADST}~\cite{lin2023discriminative}. 
Moreover, \textit{GLocalKD}~\cite{ma2022deep} and  \textit{GLADC}~\cite{luo2022deep} capture anomalies from global and local graph perspectives. Additionally, some approaches concurrently consider node-level anomalies and substructure anomalies in anomaly detection, such as \textit{iGAD}~\cite{zhang2022dual}, \textit{GLAM}~\cite{zhao2022graph} and \textit{HO-GAT}~\cite{huang2021hybrid}.

\noindent\textbf{Counterfactual Graph Learning.}
The works about counterfactual graph learning primarily fall into two folds: counterfactual explanations and counterfactual data augmentation.
The former aims to identify the necessary changes to the input graph that can alter the prediction outcome, which can help to filter out spurious
explanations~\cite{guo2023counterfactual}.
The related methods try to find a counterfactual graph by conducting minimal perturbations (i.e., adding or removing the minimum number of edges) that could lead to counterfactual predictions~\cite{bajaj2021robust,yang2023counterfactual,yang2023generating}.
Counterfactual data augmentation is a promising technique used to augment training data to reduce model reliance on spurious correlations and aid in learning causal representations~\cite{guo2023counterfactual,zhang2021learning,chen2023disco}.
To alleviate the problem of spurious correlations and enhance models' generalization capacity, 
some researchers try to inject interventions (perturbations) on the node attributes and the graph structure to generate counterfactual data~\cite{ma2022learning,xiao2023counterfactual}. While, others attend to match counterfactual examples which are the most similar items with different treatments~\cite{zhao2022learning,chang2023knowledge} to boost models' robustness.
\section{Conclusion}
\label{sec:conclusion}

In this paper, we proposed a novel framework, \modelname, for graph-level anomaly detection. 
We designed a counterfactual graph producer to produce raw counterfactual graphs by combining the discriminative motif and the contextual subgraph. It can generate high-quality counterfactual graphs and effectively alleviate the performance decline issue under varying environments.
We also proposed a GAN-based graph optimizer to refine the raw graphs through capturing the motif consistency and controlling the contextual subgraph structures. 
Extensive experiments demonstrate
the superiority of our proposed approach over state-of-the-art baselines on graph benchmarks. Our future work will focus on further improving the efficiency of \modelname~ and extending it to more complicated scenarios such as evolving graph anomaly detection.




\bibliographystyle{ACM-Reference-Format}
\bibliography{sample-base}



\end{document}